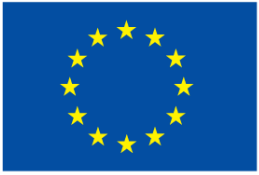 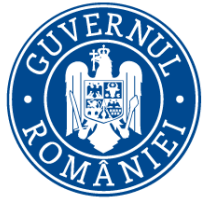 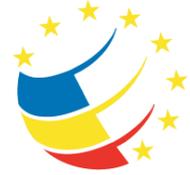

Project co-financed from the European Fund of Regional Development through the Competitivity Operational Program 2014-2020

# SpotTheFake: An Initial Report on a New CNN-Enhanced Platform for Counterfeit Goods Detection

Alexandru Șerban[1], George Ilaș[1], George-Cosmin Poruşniuc[1,2]

[1] ZentaHub SRL, Iași, Romania

[2] University of Eastern Finland, Joensuu, Finland

alexandru.serban@zentahub.com, george.ilas@zentahub.com, cosmin.porusniuc@zentahub.com

*Abstract* — The counterfeit goods trade represents nowadays more than 3.3% of the whole world trade and thus it's a problem that needs now more than ever a lot of attention and a reliable solution that would reduce the negative impact it has over the modern society. This paper presents the design and early development of a novel counterfeit goods detection platform that makes use of the outstanding learning capabilities of the classical VGG16 convolutional model trained through the process of "transfer learning" and a multi-stage fake detection procedure that proved to be not only reliable but also very robust in the experiments we have conducted so far using an image dataset of various goods we have gathered ourselves.

## I. INTRODUCTION

The counterfeit goods trade is an issue now more than ever and it's affecting not only all of the big companies producing goods for the world but also, more seriously, the end user that ultimately hopes he is paying for a genuine product and expects to receive nothing less than the quality and the outcomes that he paid for. Some of the latest studies on the matter confirm that nowadays, as much as 3.3% of the world trade is represented by illegal trafficking of counterfeit goods [1], a number that keeps rising by the day. It is a known fact that the trafficking of counterfeit products has a direct negative impact on the health and wellbeing of humanity as a whole as one of the markets most affected by this issue is the pharmaceutical industry ("World Health Organization" reported in 2017 that revenues from fake medicine made up almost 15% of the pharmaceutical market worldwide [3]). Not only this but it is also acknowledged that most (if not all) of the profit being made from the illicit trade of fake goods goes untracked by the legal authorities and this in turn affects the modern society in various hurtful ways [2].

Even though this issue has been persistent for the past few decades, it seems that both companies and authorities alike have mostly been trying to address it only through traditional means and have mostly omitted all the recent breakthroughs that have been made in the fields of computer vision and automatic recognition of objects in images. There is a wide spectrum of both overt and covert approaches that have been proposed so far for tackling the issue of detecting counterfeit products. Some examples of already in-use overt technologies include holograms, watermarks, color-shifting inks or sequential product numbering. Covert technologies are also similar to the overt ones including security inks, digital watermarks, biological, chemical or microscopic taggants, QR or RFID tags, etc. [4]. Even though all of the above-mentioned methods have already been proven to be fairly robust by how well they have been implemented in the real-world all of them come with some major downsides. Overt identification methods are heavily reliant on the presence of an authenticating detail on the surface of the object, a detail that could very easily be reverse engineered, or removed by the counterfeiters. Covert methods are usually more robust but they are not that easily adopted by the manufacturers of goods as they have to be a key process in the production scheme and some companies do not always afford to include such a delicate process in their workflow [4].

Besides the classical approaches mentioned thus far, more automated methods that are nurturing the recently-developed powers of artificial intelligence and image processing have already been proposed and successfully implemented. The best example of such a feat being achieved comes in the form of the product that **Entrupy** have been developing for the past 8 years and with which they claim to be "the only technology-based, powered by AI, authenticating solution on the market" [5]. The main idea behind their solution for the issue of luxury goods authentication is that they are using the power of **deep convolutional neural networks** along with a huge dataset of images that they have managed to put together over years and an ingenious microscopic imaging device that they are bundling with their solution and that plays the most important role in the whole detection process designed by them.

In this paper, we present the early development advances of a new platform aimed at identifying counterfeit products, that is not relying on any kind of special physical device and is instead taking full advantage of the learning capabilities of **convolutional neural networks** along with a **multi-stage detection** approach that we are proposing. The main drive behind the development of the present platform is to give more power to the end-user and enable any customer/owner of a specific product out there find out for himself whether the item that he owns is legit or not. The scope of the product is not only the one stated above but also to provide border officials or any authorities capable of fighting against the counterfeit trade business with a reliable tool that they could easily employ in their day-to-day work.

## II. RELATED WORK

Our biggest competitor out there on the market thus far seems to be **Entrupy** [5], an U.S. based company that has been actively developing their counterfeit detection solution for luxury (but not limited to) products for almost a decade. They started in 2012 with research on ways to authenticate art but have since then progressed into a very successful A.I.-driven solution for authenticating products for both buyers and sellers. The are backed up by huge names in the A.I and computer vision community such as Yann LeCun and Eric Brewer and in 2017 have published a paper [10] in which they describe the main key-points of the new counterfeit detection solution that they have brought to the market.

Firstly, they claim that the main difference they bring when comparing to classical overt/covert fake detection techniques is that with their solution, they don't have to be present during the manufacturing of the product.

Their whole solution is very reliant on wide-FOV microscopic images that they acquire with a device that produces these images with a high precision (1cm x 1cm, 200-300x magnification) and under uniform illumination.

The first supervised classifier they have tested is based on the **bag of features/bag of visual words** technique in which they use a traditional feature detector for manually extracting features from images which they then classify using a simple SVM classifier. The manage to obtain a staggering 97.8% accuracy under test settings with this first classification technique.

However, the experiments they have conducted have shown that the second classifier they employed, which is a convolutional neural network fairly similar to the VGG16 model managed to obtain better results than the first classification technique (98% test accuracy).

We believe that the main drawback related to their approach is that their solution depends upon the presence of their specialized imaging device even at the end-user level. By our proposed approach we aim to remove the necessity of such a device by only nurturing the learning capabilities of CNNs in a multi-stage detection manner.

## III. METHODS AND DATASET

### A. The data used so far

Given the nature of the considered problem, even from the conceptual phase of the development of our platform we decided that in order for our initial results to be truly meaningful we would have to come up with a way to generate our own dataset. As a result, all the training/testing data used in the experiments that will be presented next we acquired ourselves and for this purpose we designed a **data acquisition workflow** that would standardize not only the data in itself but also give a controlled setting for the assumptions we made and will make throughout the development of our solution. An overview of this flow can be observed in Fig. 1.

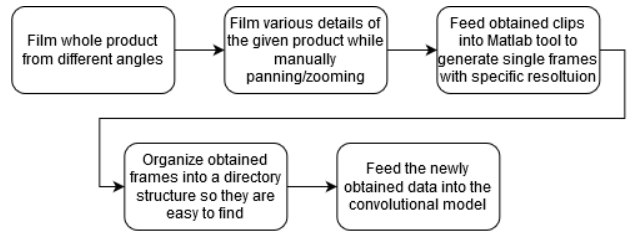

Fig. 1. The used data acquisition flow

Firstly, all the clips we filmed for generating our image data have been acquired using an iPhone X phone. We recorded short clips (5-15 seconds) of all the products chosen for initial trainings/evaluations of our approach, under roughly the same lighting conditions and using a white background. The main reason behind using this kind of context was that we don't introduce too much noise and differences in the dataset and keep the convolutional model from learning lighting information from the images during early development of our idea. Multiple short clips have been recorded from various angles of any specific product and in addition to this we also recorded clips of specific details from some of the objects (and while filming we manually panned and zoomed the image so that we would infer a certain level of invariance to the relevant features contained by the images), this having a key role in the multi-stage detection phase of our solution.

Up until the current development stage we managed to film various kinds of products such as **glasses**, **watches** and **handbags** and this we did for tackling the initial sub-problem of identifying what actual class of products are we doing the detection for. When considering the issue of **fake product detection,** we mostly focused on handbags so that we could check that the approach we adopted so far is viable and can actually render satisfying results for a specific class of products. As a result, we filmed 10 various brands of handbags along with 2 different "Louis Vuitton" bags (a small one and a larger one) for which we also filmed clips of 2 exceptionally good fake counterparts. In order to facilitate the multi-stage detection process we will shortly describe, for these 2 specific luxury bags we also recorded clips of some specific manufacturing details (buckle, logo, zipper and leather texture).

Secondly, we took all the recorded clips and fed them into a small Matlab tool which we internally developed and which would sample frames with a certain frequency from the clips and also resize the frames (224x224x3 resolution) so that they could be directly used for training the convolutional models. In Fig. 2, Fig. 3 and Fig.4, some instances of generated images from our dataset can be observed.

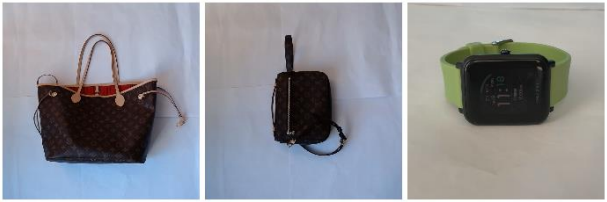

Fig. 2.   3 generated products images

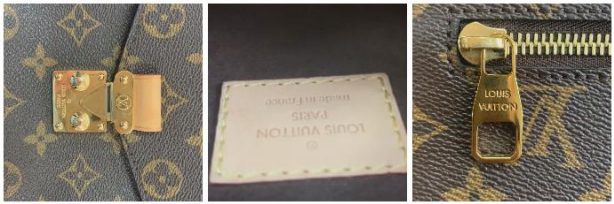

Fig. 3.   3 details from the "LV" bags

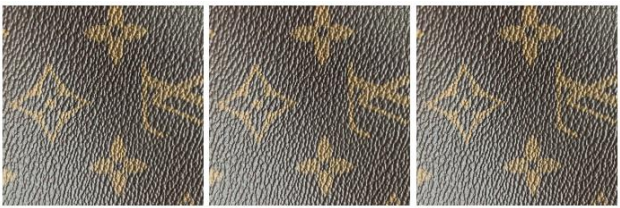

Fig. 4.   3 generated texture images

By employing the workflow described thus far we have been able to generate an initial dataset of more than 200,000 images spread across multiple classes that serve us in training both object detection models as well as fake-detection models which we will describe in the following section.

*B. The VGG16 model & Transfer Learning*

The key aspect behind the effectiveness of the method we propose lies in the already-known outstanding learning capabilities of convolutional neural networks. These are a special kind of neural networks that take full advantage of the convolution operation in their deep layers and are, as a result, capable of learning both high and low-level features from pixel data, without a specific need for any additional manual feature extraction from the data prior to it being fed into the network.

In all the experiments we've conducted thus far we used only one well established convolutional architecture, VGG16 [6], which has already been proven to work exceptionally well with data similar to what we have gathered for our solution. The VGG16 model is one of the first convolutional networks that proved the efficiency of going in deeper with the design of CNNs when regarding the task of large-scale image recognition. VGG16 gained notoriety when it managed to achieve 92.7% top-5 test accuracy [7] on ImageNet, which is a huge benchmarking dataset of over 15 million labeled high-resolution images belonging to roughly 22,000 categories [8].

The architecture in itself is quite simple and is made up of several sequentially stacked convolutional layers followed by a "head" of 2 fully connected layers. The input to the network is of fixed size (224x224x3 RGB image) and what is essential about the stack of convolutional layers is that they are using very small convolution filters of 3x3 and 1x1 (which are also seen as a linear transformation of the input channels) [7]. A more detailed overview of the classic VGG16 architecture can be observed in Fig. 5.

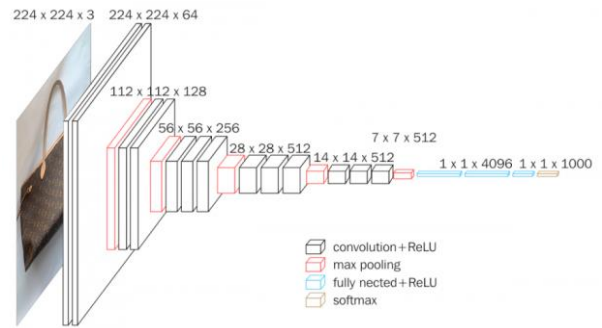

Fig. 5.   The VGG16 model. Taken from [7].

The biggest drawback of this model appears when considering training it from scratch as it requires a high amount of computational resources. That is why in our experiments we made extended use of the **transfer learning** [9] technique.

Given that it has been already well established that VGG16 performs remarkably well on the ImageNet dataset and the features in the images we have gathered thus far are very similar to the ones in some of the classes belonging to ImageNet (it contains various pictures of retail products such as bags), we figured that our solution would benefit from training a slightly modified version of VGG16 through the process of **transfer learning** so that the network would make the most use out of the already learned features from ImageNet during the pre-training phase.

As already mentioned above the first key aspect of transfer learning is that the network is not trained from scratch, but rather the training starts from a set of pre-trained weights on the ImageNet dataset (which we imported from Keras [11]). In the first stage of transfer learning, we remove the old, fully-connected head of VGG16 and replaced it with a new one containing one fully connected layer with 256 hidden units with "ReLU" activation and a new "softmax" layer with its size according to the task at hand. After the new "head" is added to the network, all the convolutional layers are firstly "frozen" and only this new head is being trained for a moderate number of iterations. Then comes the second stage of transfer learning during which all the previous convolutional layers are "unfrozen" and the network is then trained for another number of iterations. This in turn enables the network to learn the new features in the presented data, while still using the features it previously learned from the ImageNet dataset.

## C. The multi-stage detection approach

The solution we are developing will come packed as a smartphone (cross-platform) application that would work outside of the box without any other additional device being needed.

A key concept that enables us to be able to design our solution in this manner is the multi-stage fake detection approach that we want to employ. Our other competitors in the market of automatic counterfeit detection have made great progress and obtained outstanding results by heavily relying their solution on a physical device that they have developed and that they are shipping to the customer that opts for hiring their services [5]. The main drawbacks of this approach are that not only it narrows the target audience for the developed solution but also it reduces the number of usage scenarios (as one should be really careful when taking the pictures required for the authentication procedure) and it slows down the overall detection process.

To address this, our detection framework does not rely on any sort of physical device other than the smartphone camera of the user and in turn uses only the power of the VGG16 in a multi-stage detection process. Firstly, when the user wants to authenticate a product, the application would prompt him to take a picture of the whole object. That picture would be then sent to the first-stage detection model that would tell the user whether the product is already in our database our not (by *database* here we refer to whether or not our model has already seen this class of product). Secondly, if the product is recognized by the first model, the user is then prompted to take multiple photos of different details specific to that particular product. Those pictures of specific details are then sent to our second and third detection models (one takes care of authenticating whether some details like **buckles**, **labels**, **zippers** are of an original product, and the other model is more oriented only towards identifying fake/original textures of the product). Lastly, we take the prediction scores returned by the latter mentioned models and make them up into an average that would represent the final predicted percentage of originality the specific product holds. This multi-stage detection technique can be furtherly observed in the diagram shown in Fig. 6.

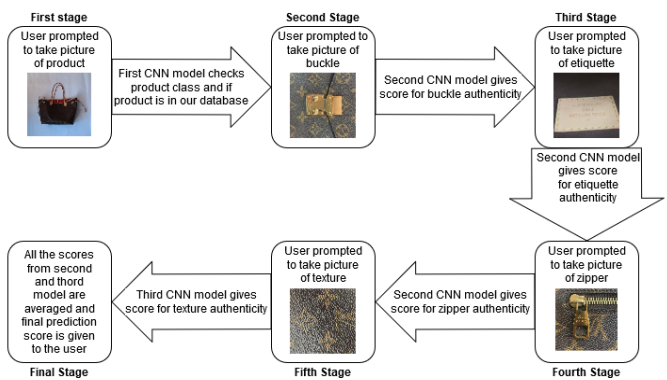

Fig. 6. An example use-case scenario for the multi-stage detection

## IV. EXPERIMENTAL RESULTS

### A. Early stage development

Throughout the development we've done so far we've managed to train various instances of the VGG16 model on different classes of data which we acquired ourselves and for benchmarking purposes we used our best 3 models so far: a model targeted towards the first, product identification stage, a model for detecting the authenticity of textures and a model for detecting the authenticity of other various specific details. After we singled out these 3 models, we deployed them on the prototype version of our platform and tested it in some real-world scenarios for detecting the authenticity of 2 "Louis Vuitton" bags on which we focused our attention thus far.

### B. Training the models

All the models we've trained thus far have the same structure described in section III.B. All the trainings have been done in a mini-batch iterative manner (with a fixed batch size of 32 images) with periodical evaluation stages being done after a specific chunk size of iterations (40-50 iterations per validation cycle, model specific). Some minor differences came when we trained the texture-specific model where we employed a slightly different approach and we also aim to further improve the texture-detection phase.

As previously mentioned, the first model we trained targets the problem of product identification. We've trained this model for a total of 4 classes which are: **glasses**, **watches**, **LV big bag** and **LV small bag.** This model has been trained with a total of 48,289 images with a validation/test split of 500 images per class (so we used 2000 images for both validation and testing after the training). The number of training iterations was 200 for both the fully-connected head training phase and the fine-tuning phase. The evolution of the training loss and accuracy for this first model can be observed in figures 7 and 8.

The second model we've trained referred the issue of specific details identification and for this purpose we also trained it using 4 classes which are: **fake_buckle**, **fake_etiquette**, **original_buckle** and **original_etiquette**. The training was conducted similarly as for the products-oriented model using a total of 11,050 images.

The third and final trained model was the one oriented towards the detection of fake textures. For training this specific model we used a slightly different approach in the sense that during training/validation/testing we tried enhancing the performance of the model by feeding 16 random smaller crops for each texture image in the training/validation batches. We needed this in the early stage development of our solution as we only managed to gather around 10,000 images of textures from the 2 different types of Louis Vuitton bags that we had (both the fake and the original ones). The evolution of training accuracy/loss for this model is presented in figures 9 and 10.

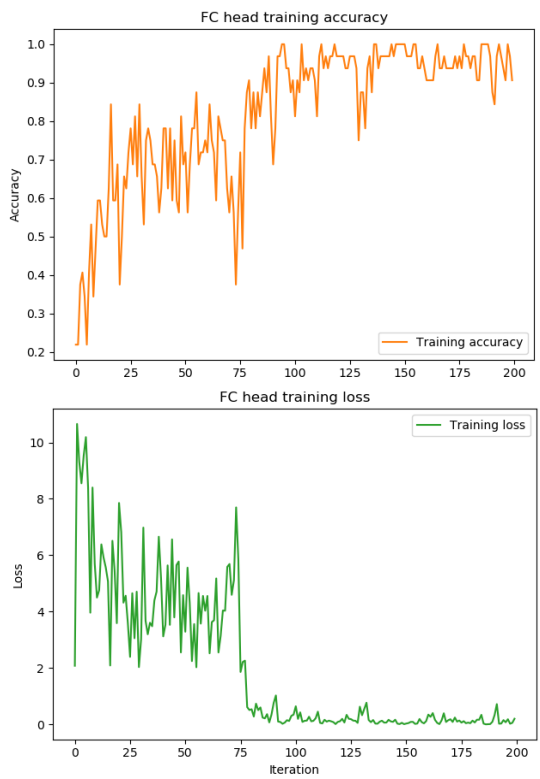

Fig. 7.  Product model fully-connected head training

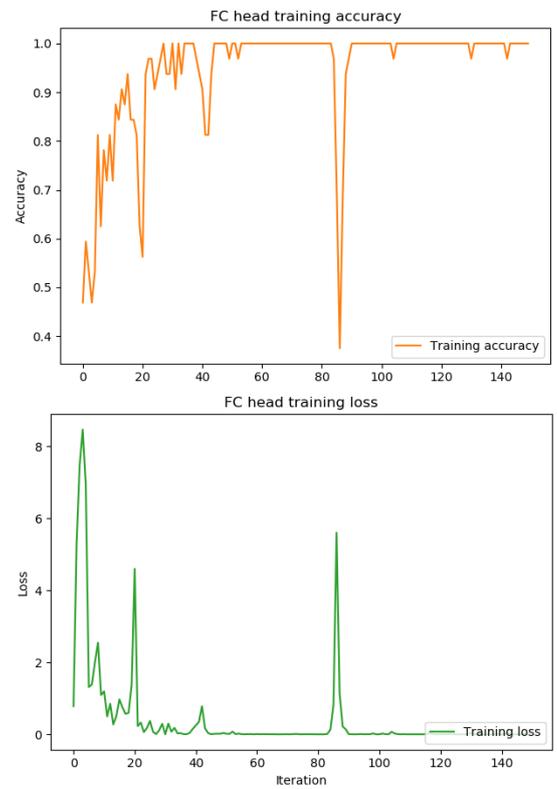

Fig. 9.  Texture model fully-connected head training

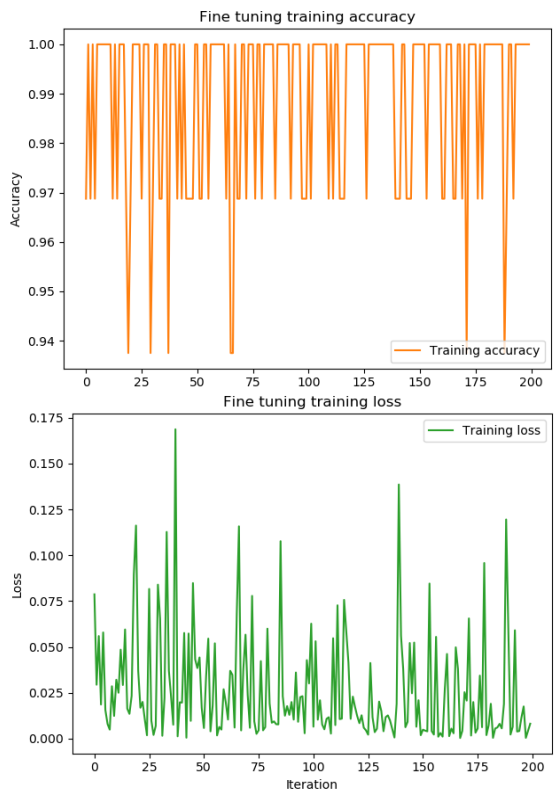

Fig. 8.  Product model fine-tuning training

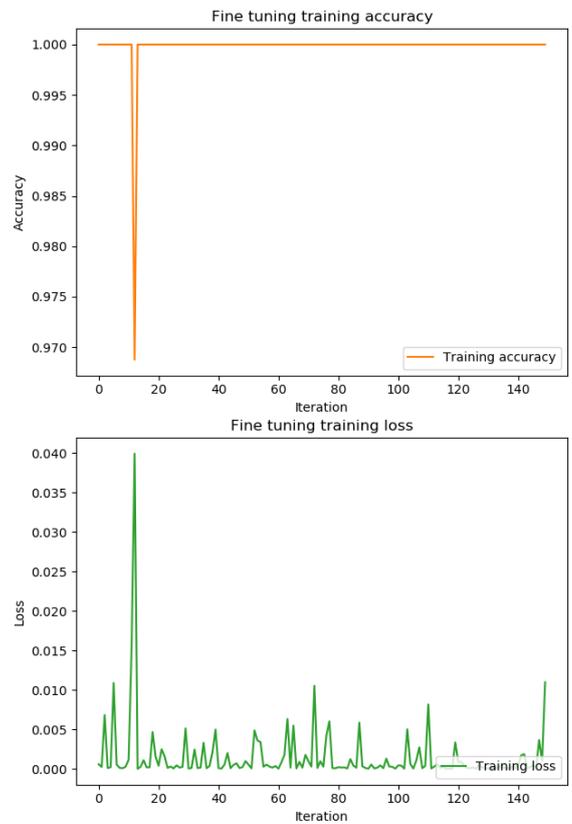

Fig. 10.  Texture model fine-tuning training

## C. Testing the models

All the training/testing of our models have been conducted on a NVIDIA RTX 2070 GPU with 8 Gb of VRAM.

When it came to testing our models and our proposed multi-stage detection method we opted for doing it both in an experimental setting (using a small subset of the generated dataset for testing purposes) and in a real-world setting, buy using our prototype platform for identifying whether our approach is robust against actually identifying whether the 2 luxury bags we've considered thus far are actually original or counterfeit (as a "real" end-user would do).

All the 3 presented models performed fantastically on the test subsets. The products model has been tested on a subset of 2000 testing images (500 for each considered class) and achieved an accuracy of almost 100% (as expected given that we used the pretrained weights on ImageNet during training). Fig 11 shows the confusion matrix of this specific model.

Both the textures and the other specific details models also performed very well on the test subsets achieving also close to 100% accuracy. The reason for this might very well be that we still need to gather more data and diversify the testing methods for our networks, but thus far these performances have been satisfactory enough for us. The confusion matrices for the other 2 trained models can be observed in figures 12 and 13.

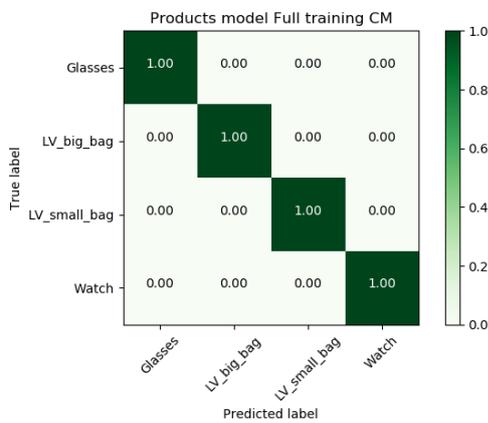

Fig. 11. Products model CM

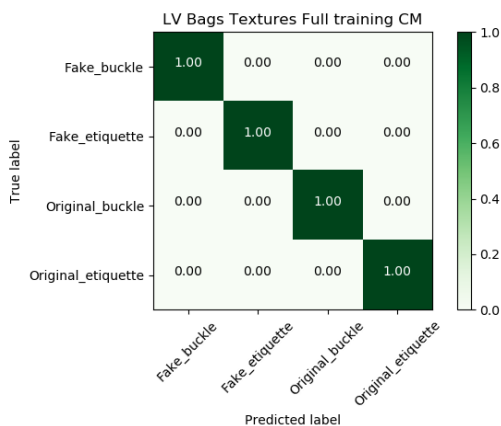

Fig. 12. Other details model CM

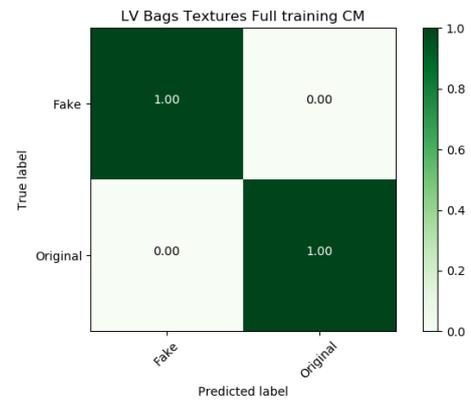

Fig. 13. Textures model CM

When we tested our solution using the application prototype that we have developed we managed to obtain average counterfeit detection scores of >90% (even though in some cases, because of the lighting and the way the data we have so far has been gathered, the scores were in the range of 60-70%).

## V. CONCLUSIONS AND FUTURE DEVELOPMENTS

Given the experiments conducted and outlined in this paper thus far we believe that the approach we propose for tackling the issue of counterfeit products identification holds a great deal of potential and we surely are on the right track to develop a robust and reliable solution for the matter in discussion that would not rely in any way or form an any kind of specific imaging hardware, enabling a broader target audience.

With respect to future developments we plan to acquire more training data by means of firstly getting a hold of more high-end products along with remarkable fakes for them and we also wish to furtherly standardize our image sampling flow by constructing a controlled environment (regarding lighting mostly, but not limited to only this aspect), where we could film the products. We could also artificially enhance the acquired images by adding artificial light sources into the photos in order to test what would be the ideal set-up for generating images from which the CNN models would manage to extract the best features possible.

We plan on furtherly improving and developing the detection of fakes regarding the textures of the products and in this way, we will look at various different convolutional models along with new ways of pre-processing the texture images so that they would be enough for developing a robust solution without the need of microscopic imaging.